\documentclass{article}

\usepackage[preprint, nonatbib]{neurips_2020}

\usepackage[utf8]{inputenc} %
\usepackage[T1]{fontenc}    %
\usepackage{hyperref}       %
\usepackage{url}            %
\usepackage{booktabs}       %
\usepackage{amsfonts}       %
\usepackage{nicefrac}       %
\usepackage{marvosym}       %
\usepackage{microtype}      %
\usepackage{xcolor}
\usepackage{xspace}
\usepackage{textcomp} %
\usepackage{listings}

\definecolor{codegreen}{rgb}{0,0.6,0}
\definecolor{codegray}{rgb}{0.5,0.5,0.5}
\definecolor{codepurple}{rgb}{0.58,0,0.82}
\definecolor{backcolour}{rgb}{0.95,0.95,0.92}

\lstdefinestyle{mystyle}{
    backgroundcolor=\color{backcolour},   
    commentstyle=\color{codegreen},
    keywordstyle=\color{magenta},
    numberstyle=\tiny\color{codegray},
    stringstyle=\color{codepurple},
    basicstyle=\ttfamily\footnotesize,
    breakatwhitespace=false,         
    breaklines=true,                 
    captionpos=b,                    
    keepspaces=true,                 
    numbers=left,                    
    numbersep=5pt,                  
    showspaces=false,                
    showstringspaces=false,
    showtabs=false,                  
    tabsize=2,
    upquote=true
}
\hypersetup{hidelinks} %

    \usepackage{adjustbox, array, booktabs, multirow}
    
    \newcolumntype{R}[2]{%
        >{\adjustbox{angle=#1,lap=\width-(#2)}\bgroup}%
        l%
        <{\egroup}%
    }
    \newcommand*\rot{\multicolumn{1}{R{90}{-0.5em}}}%

    \newcommand*\rotBells{\multicolumn{1}{R{90}{-0.8em}}}%

\newcommand{\todo}[1]{}
\newcommand{\Al}[1]{}
\newcommand{\Fo}[1]{}
\newcommand{\Ra}[1]{}
\newcommand{\atodo}[1]{}

\newcommand{\Alnew}[1]{}

\newcommand{\etal}{{\em et~al.}\xspace}

\newcommand{\vsection}[1]{\section{#1}\vspace{-0.3cm}}
\newcommand{\vsubsection}[1]{\subsection{#1}\vspace{-0.2cm}}
\newcommand{\vsubsubsection}[1]{\subsubsection{#1}\vspace{-0cm}}
\newcommand{\vparagraph}[1]{\vspace{-0cm}\paragraph{#1}}

\title{SqueezeBERT: What can computer vision teach NLP about efficient neural networks?}

\author{%
    Forrest N. Iandola \\
    {\scriptsize \texttt{forresti@berkeley.edu}} \\
    \And
    Albert E. Shaw \\
    {\scriptsize \texttt{ashaw596@gmail.com}} \\
    \And
    Ravi Krishna\\
    UC Berkeley EECS \\
    {\scriptsize \texttt{ravi.krishna@berkeley.edu}}
    \And
    Kurt W. Keutzer \\
    UC Berkeley EECS \\
    {\scriptsize \texttt{keutzer@berkeley.edu}}
}

\begin{document}
\maketitle

\setcounter{footnote}{0} %

\begin{abstract}
Humans read and write hundreds of billions of messages every day.
Further, due to the availability of large datasets, large computing systems, and better neural network models, natural language processing (NLP) technology has made significant strides in understanding, proofreading, and organizing these messages.
Thus, there is a significant opportunity to deploy NLP in myriad applications to help web users, social networks, and businesses.
In particular, we consider smartphones and other mobile devices as crucial platforms for deploying NLP models at scale.
However, today's highly-accurate NLP neural network models such as BERT and RoBERTa are extremely computationally expensive, with BERT-base taking 1.7 seconds to classify a text snippet on a Pixel 3 smartphone.
In this work, we observe that methods such as grouped convolutions have yielded significant speedups for computer vision networks, but many of these techniques have not been adopted by NLP neural network designers.
We demonstrate how to replace several operations in self-attention layers with grouped convolutions, and we use this technique in a novel network architecture called SqueezeBERT, which runs 4.3x faster than BERT-base on the Pixel 3 while achieving competitive accuracy on the GLUE test set.
The SqueezeBERT code will be released.

\end{abstract}

\vsection{Introduction and Motivation}
\label{sec_intro}
The human race writes over 300 billion messages per day~\cite{2019_twitter_per_day, 2019_fb_per_day, 2018_whatsapp_per_day, 2017_email_per_day}.
Out of these, more than half of the world's emails are read on mobile devices, and nearly half of Facebook users exclusively access Facebook from a mobile device~\cite{2017_email_mobile, 2018_facebook_mobile}.
Natural language processing (NLP) technology has the potential to aid these users and communities in several ways.
When a person writes a message, NLP models can help with spelling and grammar checking as well as sentence completion. When content is added to a social network, NLP can facilitate content moderation before it appears in other users' news feeds. When a person consumes messages, NLP models can help classify messages into folders, composing news feeds, prioritizing messages, and identifying duplicate messages.

In recent years, the development and adoption of Attention Neural Networks has led to dramatic improvements in almost every area of NLP.
In 2017, Vaswani~\etal proposed the multi-head self-attention module, which demonstrated superior accuracy to recurrent neural networks on English-German machine language translation~\cite{2017_attention_vaswani}.\footnote{Neural networks that use the self-attention modules of Vaswani~\etal are sometimes called "Transformers," but in the interest of clarity we call them "self-attention networks."}
These modules have since been adopted by GPT~\cite{2018_gpt} and BERT~\cite{2019_BERT} for sentence classification, and by GPT-2~\cite{2019_gpt2} and CTRL~\cite{2019_ctrl} for sentence completion and generation. Recent works such as ELECTRA~\cite{2020_electra} and RoBERTa~\cite{2019_roberta} have shown that larger datasets and more sophisticated training regimes can further improve the accuracy of self-attention networks.

Considering the enormity of the textual data created by humans on mobile devices, a natural approach is to deploy NLP models themselves on the mobile devices themselves, embedding them in common apps that are used to read, write, and share text.
Unfortunately, many of today's best state-of-the-art NLP models may are often rather computationally expensive, often making mobile deployment impractical.
For example, we observe that running the BERT-base network on a Google Pixel 3 smartphone approximately 1.7 seconds to classify a single text data sample.\footnote{Note that BERT-base~\cite{2019_BERT}, RoBERTa-base~\cite{2019_roberta}, and ELECTRA-base~\cite{2020_electra} all use the same self-attention encoder architecture, and therefore these networks incur approximately the same latency on a smartphone.} Much of the research on efficient self-attention networks for NLP has just emerged in the past year. However, starting with SqueezeNet~\cite{2016_squeezenet}, the mobile computer vision (CV) community has spent the last four years optimizing neural networks for mobile devices. 
Intuitively, it seems like there must be opportunities to apply the lessons learned from the rich literature of mobile CV research to accelerate mobile NLP. In the following we review what has already been applied and propose two additional techniques from CV that we will leverage for accelerating NLP.

\vsubsection{What has CV research already taught NLP research about efficient networks?}
In recent months, novel self-attention networks have been developed with the goal of achieving faster inference.
At present, the MobileBERT network defines the state-of-the-art in low-latency text classification for mobile devices~\cite{2020_MobileBERT_ACL}.
MobileBERT takes approximately 0.6 seconds to classify a text sequence on a Google Pixel 3 smartphone.
And, on the GLUE benchmark, which consists of 9 natural language understanding (NLU) datasets~\cite{2018_glue}, MobileBERT achieves higher accuracy than other efficient networks such as DistilBERT~\cite{2019_distilbert}, PKD~\cite{2019_pkd}, and several others~\cite{2019_albert, 2019_well_read, 2019_tinybert, 2020_theseus}.
To achieve this, MobileBERT introduced two concepts into their NLP self-attention network that are already in widespread use in CV neural networks:
\begin{enumerate}
      \item {\bf Bottleneck layers.} In ResNet~\cite{2015_resnet}, the 3x3 convolutions are computationally expensive, so a 1x1 "bottleneck" convolution is employed to reduce the number of channels input to each 3x3 convolution layer. Similarly, MobileBERT adopts bottleneck layers that reduce the number of channels before each self-attention layer, and this reduces the computational cost of the self-attention layers.
      \item {\bf High-information flow residual connections.} In BERT-base, the residual connections serve as links between the low-channel-count (768 channels) layers. The high-channel-count (3072 channels) layers in BERT-base do not have residual connections. However, the ResNet and Residual-SqueezeNet~\cite{2016_squeezenet} CV networks connect the high-channel-count layers with residuals, which enables higher information flow through the network. Similar to these CV networks, MobileBERT adds residual connections between the high-channel-count layers.
\end{enumerate}

\vsubsection{What else can CV research teach NLP research about efficient networks?}
We are encouraged by the progress that MobileBERT has made in leveraging ideas that are popular in the CV literature to accelerate NLP. 

However, we are aware of two other ideas from CV, which weren't used in MobileBERT, and that could be applied to accelerate NLP:
\begin{enumerate}
     \item {\bf Convolutions.} Since the 1980s, computer vision neural nets have relied heavily on convolutional layers~\cite{1980_conv, 1989_conv}. Convolutions are quite flexible and well-optimized in software, and they can implement things as simple as a 1D fully-connected layer, or as complex as a 3D dilated layer that performs upsampling or downsampling.
     \item {\bf Grouped convolutions.} A popular technique in modern mobile-optimized neural networks is grouped convolutions (see Section \ref{sec_grouped_conv}). Proposed by Krizhevsky \etal in the 2012 winning submission to the ImageNet image classification challenge \cite{2011_cuda_convnet,2012_alexnet, 2015_imagenet}, grouped convolutions disappeared from the literature from some years, then re-emerged as a key technique circa 2016~\cite{2016_xception, 2016_resnext} and today are extensively used in efficient CV networks such as MobileNet~\cite{2017_mobilenets}, ShuffleNet~\cite{2018_shufflenet}, and EfficientNet~\cite{2019_efficientnet}. While common in efficient CV literature, we are not aware of work applying grouped convolutions to NLP.
\end{enumerate}

\vsubsection{SqueezeBERT: Applying lessons learned from CV to NLP}
In this work, we describe how to apply convolutions and particularly grouped convolutions in the design of a novel self-attention network for NLP, which we call SqueezeBERT. Empirically, we find that SqueezeBERT runs at lower latency on a smartphone than BERT-base, MobileBERT, and several other efficient NLP models, while maintaining competitive accuracy.

\vsection{Implementing self-attention with convolutions}
\label{sec_conv}
In this section, first we review the basic structure of self-attention networks, next we identify that their biggest computational bottleneck is in their position-wise fully-connected (PFC) layers, and then we show that these PFC layers are equivalent to a 1D convolution with a kernel size of 1.

\vsubsection{Self-attention networks}

In most BERT-derived networks there are typically 3 stages: the embedding, the encoder, and the classifier~\cite{2019_BERT, 2019_roberta, 2020_electra, 2020_MobileBERT_ACL, 2019_albert}.
\footnote{Some self-attention networks such as~\cite{2017_attention_vaswani, 2018_gpt} also have "decoder'' stage. The decoder typically uses a similar neural architecture as the encoder, but is auto-regressive.}
The embedding converts preprocessed words (represented as integer-valued tokens) into learned feature-vectors of floating-point numbers; the encoder is comprised of a series of self-attention and other layers; and the classifier produces the network's final output.
As we will see later in Table~\ref{T_bert_compute_breakdown}, the embedding and the classifier account for less than 1\% of the runtime of a self-attention network, so we focus our discussion on the encoder.

We now describe the encoder that is used in BERT-base~\cite{2019_BERT}.
The encoder consists of a stack of blocks.
Each block consists of a self-attention module followed by three position-wise fully-connected layers, known as feed-forward network (FFN) layers.
Each self-attention module contains three seperate position-wise fully-connected (PFC) layers, which are used to generate the \emph{query} ($Q$), \emph{key} ($K$), and \emph{value} ($V$) activation vectors for each position in the feature embedding. 
Each of these PFC layers in self-attention applies the same operation to each position in the feature embedding independently.
While neural networks traditionally multiply weights by activations, a distinguishing factor of attention neural networks is that they multiply activations by other activations, which enables dynamic weighting of tensor elements to adjust based on the input data.
Further, attention networks allow modeling of arbitrary dependencies irregardless of their distance in the input or output~\cite{2017_attention_vaswani}.
The self-attention module proposed by Vaswani \etal~\cite{2017_attention_vaswani} (which is also used by GPT~\cite{2018_gpt}, BERT~\cite{2019_BERT}, RoBERTa~\cite{2019_roberta}, ELECTRA~\cite{2020_electra} and others) multiplies the $Q$, $K$, and $V$ activations together using the equation $softmax(\frac{QK^T}{\sqrt{d_k}})V$, where $d_k$ is the number of channels in one attention head.\footnote{For example, in BERT-base, the self-attention module has 768 channels and 12 heads, so $d_k = \frac{768}{12} = 64$.}

\begin{table*}[t!]
    \caption{{\bf How does BERT spend its time?} This is a breakdown of computation (in floating-point operations, or FLOPs) and latency (on a Google Pixel 3 smartphone) in BERT-base, reported to three significant digits. The FC layers account for more than 97\% of the FLOPs and over 88\% of the latency.}
\vspace{1mm}
    \centering
    \begin{tabular}{llcc}
      \toprule
      Stage & Module type & FLOPs & Latency \\
      \midrule
      Embedding & Embedding & 0.00\% & 0.26\% \\
      Encoder & FC in self-attention modules & 24.3\% & 18.9\% \\
      Encoder & $softmax(\frac{QK^T}{\sqrt{d_k}})V$ in self-attention modules & 2.70\% & 11.3\% \\
      Encoder & FC in feed-forward network layers & 73.0\% & 69.4\% \\
      Final Classifier & FC layers in final classifier & 0.00\% & 0.02\% \\
      \midrule
      Total &  & 100\% & 100\% \\
      \bottomrule
    \end{tabular}
    \label{T_bert_compute_breakdown}
    \vspace{2mm}
\end{table*}

\vsubsection{Benchmarking BERT for mobile inference}
To identify the parts of BERT that are time-consuming to compute, we profile BERT on a smartphone.
Specifically, we measure the neural network's latency using PyTorch~\cite{2019_pytorch} and TorchScript on a Google Pixel 3 smartphone, with an input sequence length of 128 and a batch size of 1. %
In Table~\ref{T_bert_compute_breakdown}, we show the breakdown of FLOPs and latency among the main components of the BERT network, and we observe that the $softmax(\frac{QK^T}{\sqrt{d_k}})V$ calculations in the self-attention modules account for only 11.3 percent of the total latency.
However, the PFC layers in the self-attention modules account for 18.9 percent, and the PFC layers in the feed-forward network modules account for 69.4 percent, and in total the PFC layers account for 88.3 percent of the latency.
Given that the PFC layers account for the overwhelming majority of the latency, we now turn our focus to reducing the latency of the PFC layers.

\vsubsection{Replacing the position-wise fully connected layers (PFC) with convolutions}
To address this, we intend to replace the PFC layers with grouped convolutions, which have been shown to produce significant speedups in computer vision networks.
As a first step in this direction, we now show that the fully-connected layers used throughout the BERT encoder are a special case of non-grouped 1D convolution.
In the following, $f$ denotes the input feature vector, and $w$ denotes the weights.
Given an input feature vector of dimensions $(P,C_{in})$ with $P$ positions and $C_{in}$ channels to generate an output of $(P,C_{out})$ features, the operation performed by the position-wise fully-connected layer can be defined as follows:
\begin{equation}
    PositionwiseFullyConnected_{p,c_{out}}(f,w) = {\sum^{C_{in}}_i{w_{c_{out},i}*f_{p,i}}}
\end{equation}

Then if we consider the definition of a 1D convolution with kernel size $K$ with the same input and output dimensions
\begin{equation}
    Convolution_{p,c_{out}}(f,w) = {\sum^{C_{in}}_i\sum^K_k{{w_{c_{out},i,k}*f_{(p-\frac{K-1}{2}+k),i}}}}
\end{equation}

we observe that the position-wise fully-connected operation is equivalent to a convolution with a kernel size of $k=1$.
Thus, the PFC layers of Vaswani~\etal~\cite{2017_attention_vaswani}, GPT, BERT, and similar self-attention networks can be implemented using convolutions without changing the networks' numerical properties or behavior.

\vsection{Incorporating grouped convolutions into self-attention}
\label{sec_grouped_conv}
Now that we have shown how to implement the expensive PFC layers in self-attention networks using convolutions, we can incorporate efficient grouped convolutions into a self-attention network.
Grouped convolutions are defined as follows.
Given an input feature vector of dimensions $(P,C_{in})$ with P positions and $C_{in}$ channels and outputting a vector with dimensions $(P,C_{out})$,
a 1d convolution with kernel size $K$ and $G$ groups can be defined as follows.
\begin{equation}
    GroupedConvolution_{p,c_{out}}(f,w) = {\sum^{\frac{C_{in}}{G}}_{i}\sum^K_k{{w_{c_{out},i,k}*f_{(p-\frac{K-1}{2}+k), (i + \left\lfloor \frac{(i)( G)}{C_{out}}\right\rfloor \frac{C_{in}}{G})}}}}
\end{equation}

This is equivalent to splitting the the input vector into $G$ separate vectors of size $(P,\frac{C_{in}}{G})$ along the $P$ dimension and running $G$ separate convolutions with independent weights each computing vectors of size $(P, \frac{C_{out}}{G})$.
The grouped convolution, however, requires only $\frac{1}{G}$ as many floating point operations (FLOPs) and $\frac{1}{G}$ as many weights as an ordinary convolution, not counting the small (and unchanged) amount of operations needed for the channel-wise bias term that is often included in convolutional layers.\footnote{Note that the grouped convolution with $G=1$ is identical to an ordinary convolution.}

\vsubsection{SqueezeBERT}
Now, we describe our proposed neural architecture called SqueezeBERT, which uses grouped convolutions.
SqueezeBERT is much like BERT-base, but with PFC layers implemented as convolutions, and grouped convolutions for many of the layers.
Recall from Section~\ref{sec_conv} that each block in the BERT-base encoder has a self-attention module with 3 PFC layers, plus 3 more PFC layers called feed-forward network layers (FFN$_1$, FFN$_2$, and FFN$_3$).
The FFN layers have the following dimensions: FFN$_1$ has $C_{in}=C_{out}=768$, FFN$_2$ has $C_{in}=768$ and $C_{out}=3072$, and FFN$_3$ has $C_{in}=3072$ and $C_{out}=768$.
In all PFC layers of the self-attention modules, and in the FFN$_2$ and FFN$_3$ layers, we use grouped convolutions with $G=4$.
To allow for mixing across channels of different groups, we use $G=1$ in the less-expensive FFN$_1$ layers.
Note that in BERT-base, FFN$_2$ and FFN$_3$ each have 4 times more arithmetic operations than FFN$_1$.
However, when we use $G=4$ in FFN$_2$ and FFN$_3$, now all FFN layers have the same number of arithmetic operations.

Finally, the embedding size (768), the number of blocks in the encoder (12), the number of heads per self-attention module (12), the tokenizer (WordPiece~\cite{2012_wordpiece,2016_wordpiece}), and other aspects of SqueezeBERT are adopted from BERT-base.
Aside from the convolution-based implementation and the adoption of grouped convolutions, the SqueezeBERT architecture is identical to BERT-base.

\vsection{Experimental Methodology}

\vsubsection{Datasets}
\vparagraph{Pretraining Data.} For pretraining, we use a combination of Wikipedia and BooksCorpus \cite{2015_bookscorpus}, setting aside 3\% of the combined dataset as a test set. Following the ALBERT paper, we use Masked Language Modeling (MLM) and Sentence Order Prediction (SOP) as pretraining tasks~\cite{2019_albert}.

\vparagraph{Finetuning Data.} We finetune and evaluate SqueezeBERT (and other baselines) on the General Language Understanding Evaluation (GLUE) set of tasks. This benchmark consists of a diverse set of 9 NLU tasks; thanks to the structure and breadth of these tasks (see supplementary material for detailed task-level information), GLUE has become the standard evaluation benchmark for NLP research. A model's performance across the GLUE tasks likely provides a good approximation of that model's generalizability (esp. to other text classification tasks).

\vsubsection{Training Methodology}
Many of the recent papers on efficient NLP networks report results on models trained with bells and whistles such as distillation, adversarial training, and/or transfer learning across GLUE tasks.
However, there is no standardization of these training schemes across different papers, making it difficult to distinguish the contribution of the model from the contribution of the training scheme to the final accuracy number.
Therefore, we first train SqueezeBERT using a simple training scheme (described in Section~\ref{sec_training_without_bells}, with results reported in Section~\ref{sec_results_without_bells}), and then we train SqueezeBERT with distillation and other techniques (described in Section~\ref{sec_training_with_bells}, with results reported in Section~\ref{sec_results_with_bells}).

\vsubsubsection{Training without bells and whistles}
\label{sec_training_without_bells}
We pretrain SqueezeBERT from scratch (without distillation) using the LAMB optimizer, and we employ the hyperparameters recommended by the LAMB authors: a global batch size of 8192, a learning rate of 2.5e-3, and a warmup proportion of 0.28~\cite{2020_lamb}.
Following the LAMB paper's recommendations, we pretrain for 56k steps with a maximum sequence length of 128 and then for 6k steps with a maximum sequence length of 512.

For finetuning, we use the AdamW optimizer with a batch size of 16 without momentum or weight decay with $\beta_1=0.9$ and $\beta_2=0.999$~\cite{2019_adamw}.
As is common in the literature, during finetuning for each task, we perform hyperparameter tuning on the learning rate and dropout rate. We present more details on this in the supplementary material.
In the interest of a fair comparison, we also train BERT-base using the aforementioned pretraining and finetuning protocol.

\vsubsubsection{Training with bells and whistles}
\label{sec_training_with_bells}
We now review recent techniques for improving the training of NLP networks, and we describe the approaches that we will use for the training and evaluation of SqueezeBERT in Section~\ref{sec_results_with_bells}.

\vparagraph{Distillation approaches used in other efficient NLP networks.}
While the term "knowledge distillation" was coined by Hinton~\etal to describe a specific method and equation~\cite{2015_distillation}, the term "distillation" is now used in reference to a diverse range of approaches where a "student" network is trained to replicate a "teacher" network.
Some researchers distill only the final layer of the network~\cite{2019_distilbert}, while others also distill the hidden layers~\cite{2020_MobileBERT_ACL, 2019_pkd, 2020_theseus}.
When distilling the hidden layers, some apply layer-by-layer distillation warmup, where each module of the student network is distilled independently while downstream modules are frozen~\cite{2020_MobileBERT_ACL}.
Some distill during pretraining~\cite{2020_MobileBERT_ACL, 2019_distilbert}, some distill during finetuning~\cite{2020_theseus}, and some do both~\cite{2019_pkd, 2019_tinybert}.

\vparagraph{Bells and whistles used for training SqueezeBERT (for results in Section~\ref{sec_results_with_bells}).}
Distillation is not a central focus of this paper, and there is a large design space of potential approaches to distillation, so we select a relatively simple form of distillation for use in SqueezeBERT training.
We apply distillation only to the final layer, and only during finetuning.
On the GLUE sentence classification tasks, we use soft cross entropy loss with respect to a weighted sum of the teacher's logits and a one-hot encoding of the ground-truth.
Also note that GLUE has one regression task (STS-B text similarity), and for this task we replace the soft cross entropy loss with mean squared error.
In addition to distillation, inspired by STILTS~\cite{2018_stilts} and ELECTRA~\cite{2020_electra} we apply transfer learning from the MNLI GLUE task to other GLUE tasks as follows.
The SqueezeBERT student model is pretrained using the approach described in Section~\ref{sec_training_without_bells}, and then it is finetuned on the MNLI task.
The weights from MNLI training are used as the initial student weights for other GLUE tasks except for CoLA.\footnote{For CoLA, the student weights are pretrained (per Section~\ref{sec_training_without_bells}) but not finetuned on MNLI prior to task-specific training.}
Similarly, the teacher model is a BERT-base model that is pretrained using the ELECTRA method and then finetuned on MNLI.
Then, the teacher model is finetuned independently on each GLUE task, and these task-specific teacher weights are used for distillation.

\begin{table*} \centering
    \caption{Comparison of neural networks on the {\bf development} set of the GLUE benchmark. \Cross denotes models trained by the authors of the present paper. Bells and whistles are: {\em A} = adversarial training; {\em D} = distillation of final layer; {\em E} = distillation of encoder layers; {\em S} = transfer learning across GLUE tasks (a.k.a. STILTs~\cite{2018_stilts}); {\em W} = per-layer warmup. In GLUE accuracy, a dash means that accuracy for this task is not provided in the literature.}

    \footnotesize
    \centering
    \begin{tabular}{p{2.3cm}c|cccccccccc|cccc}
        \toprule
        & & \multicolumn{10}{c}{GLUE accuracy} & \multicolumn{4}{|c}{efficiency}  \\
        \midrule
        Model & \rotBells{Bells \& Whistles} & \rot{MNLI-m} & \rot{MNLI-mm} & \rot{QQP} & \rot{QNLI} & \rot{SST-2} & \rot{CoLA} & \rot{STS-B} & \rot{MRPC} & \rot{RTE} & \rot{Average} & \rot{\#MParams} & \rot{GFLOPs} & \rot{Latency (ms)} & \rot{Speedup} \\
        \midrule
        & & \multicolumn{10}{l|}{\em Results {\bf \em without} bells and whistles} \\
        BERT-base\Cross                                & - & 85.2 & 84.8 & 89.9 & 92.2 & 92.7 & 62.8 & 90.7 & 91.2 & 76.5 & 85.1 & 109  & 22.5 & 1690 & 1.0x \\ %
        MobileBERT~\cite{2020_MobileBERT_ACL}          & - & 80.8 & -    & -    & 88.2 & 90.1 & -    & -    & 84.3 & -    & -    & 25.3 & 5.36 & 572  & 3.0x \\
        {\scriptsize ALBERT-base~\cite{2019_albert}}   & - & 81.6 & -    & -    & -    & 90.3 & -    & -    & -    & -    & -    & 12.0 & 22.5 & 1690 & 1.0x  \\
        SqueezeBERT\Cross                              & - & 82.3 & 82.9 & 89.4 & 90.5 & 92.0 & 53.7 & 89.4 & 89.8 & 71.8 & 82.4 & 51.1 & 7.42 & 390  & 4.3x \\
        \midrule
        & & \multicolumn{10}{l|}{\em Results {\bf \em with} bells and whistles} \\
        {\scriptsize DistilBERT~6/768~\cite{2019_distilbert}}    & D   & 82.2 & -     & 88.5 & 89.2 & 91.3 & 51.3 & 86.9 & 87.5 & 59.9 & -   & 66 & 11.3 & 814 & 2.1x   \\ %
        Turc~6/768~\cite{2019_well_read}             & D   & 82.5 & 83.4 & 89.6 & 89.4 & 91.1 & -  & -  & 87.2 &	66.7 & - & 67.5 & 11.3 & 814 & 2.1x        \\ %
        Theseus~6/768~\cite{2020_theseus}             & DESW & 82.3 & -  &  89.6 & 89.5 & 91.5 & 51.1 & 88.7 & 89.0 & 68.2 & - & 66 & 11.3 & 814 & 2.1x  \\ %
        MobileBERT~\cite{2020_MobileBERT_ACL}          & DEW & 84.4 & -    & -    & 91.5 & 92.5 & -    & -    & 87.0 & -    & -    & 25.3 & 5.36 & 572  & 3.0x \\
        SqueezeBERT\Cross                              & DS & 82.5 & 82.9 & 89.5 & 90.9 & 92.2 & 53.7 & 90.3 & 92.0 & 80.9 & 84.0 & 51.1 & 7.42 & 390  & 4.3x \\
        \bottomrule
    \end{tabular}
    \label{T_dev_set}

\end{table*}

\begin{table*} \centering
    \caption{Comparison of neural networks on the {\bf test} set of the GLUE benchmark. \Cross denotes models trained by the authors of the present paper. Bells and whistles are: {\em A} = adversarial training; {\em D} = distillation of final layer; {\em E} = distillation of encoder layers; {\em S} = transfer learning across GLUE tasks (a.k.a. STILTs~\cite{2018_stilts}); {\em W} = per-layer warmup.}

    \footnotesize
    \centering
    \begin{tabular}{p{2.3cm}c|ccccccccccc|cccc}
        \toprule
        & & \multicolumn{11}{c}{GLUE accuracy} & \multicolumn{4}{|c}{efficiency}  \\
        \midrule
        Model  & \rotBells{Bells \& Whistles} & \rot{MNLI-m} & \rot{MNLI-mm} & \rot{QQP} & \rot{QNLI} & \rot{SST-2} & \rot{CoLA} & \rot{STS-B} & \rot{MRPC} & \rot{RTE} & \rot{WNLI} & \rot{GLUE score} & \rot{\#MParams} & \rot{GFLOPs} & \rot{Latency (ms)} & \rot{Speedup} \\
        \midrule
        & & \multicolumn{11}{l|}{\em Results {\bf \em without} bells and whistles} \\
        BERT-base\Cross                               & -   & 84.4 & 84.2 & 80.5 & 91.4 & 92.8 & 51.3 & 86.9 & 87.9 & 70.7 & 65.1 & 79.0 & 109  & 22.5 & 1690 & 1.0x \\ %
        BERT-base~\cite{2019_BERT}                    & -   & 84.6 & 83.4 & 80.2 & 90.5 & 93.5 & 52.1 & 86.5 & 86.9 & 66.4 & 65.1 & 78.3 & 109  & 22.5 & 1690 & 1.0x \\ %
        SqueezeBERT\Cross                             & -   & 82.0 & 81.1 & 80.1 & 90.1 & 91.0 & 46.5 & 84.9 & 86.1 & 66.7 & 65.1 & 76.9 & 51.1 & 7.42 & 390  & 4.3x \\
        \midrule
        & & \multicolumn{11}{l|}{\em Results {\bf \em with} bells and whistles} \\
        {\scriptsize TinyBERT~4/312~\cite{2019_tinybert}}  & DE  & 82.5 & 81.8 & - & 87.7 & 92.6 & 43.3 & 79.9 & - & 62.9 & 65.1 & - & 14.5 & 1.2 & 118 & 14x \\
        {\tiny ELECTRA-Small++~\cite{2020_electra}}    & AS  & 81.6 & - & - & 88.3 & 91.1 & 55.6 & 84.6 & 84.9 & 63.6 & 65.1 & - & 14.0 & 2.62 & 248 & 6.8x \\ %
        PKD~6/768~\cite{2019_pkd}                      & DE  & 81.5 & 81.0 & 79.8 & 89.0 & 92.0 & -   & -   & 82.5 & -    & 65.1 & - & 67.0 & 11.3 & 814 & 2.1x \\ %
        Turc~6/768~\cite{2019_well_read}               & D   & 82.8 & 82.2 & 79.7 & 89.4 & 91.8 & -   & -   & 84.3 & 65.3 & 65.1 & - & 67.5 & 11.3 & 814 & 2.1x  \\ %
        Theseus~6/768~\cite{2020_theseus}              & DESW& 82.4 & 82.1 & 80.5 & 89.6 & 92.2 & 47.8 & 84.9 & 85.4 & 66.2 & 65.1 & 77.1 & 66 & 11.3 & 814 & 2.1x \\
        MobileBERT~\cite{2020_MobileBERT_ACL}          & DEW & 84.3 & 83.4 & 79.4 & 91.6 & 92.6 & 51.1 & 85.5 & 86.7 & 70.4 & 65.1 & 78.5 & 25.3 & 5.36 & 572  & 3.0x \\     %
        SqueezeBERT\Cross                              & DS  & 82.0 & 81.1 & 80.3 & 90.1 & 91.4 & 46.5 & 86.7 & 87.8 & 73.2 & 65.1 & 78.1 & 51.1 & 7.42 & 390  & 4.3x \\

        \bottomrule
    \end{tabular}
    \label{T_test_set}
\vspace{-1mm}
\end{table*}

\vsection{Results}
We now turn our attention to comparing SqueezeBERT to other efficient neural networks.

\vsubsection{Results {\em without} bells and whistles}
\label{sec_results_without_bells}

In the upper portions of Tables~\ref{T_dev_set} and~\ref{T_test_set}, we compare our results to other efficient networks on the dev and test sets of the GLUE benchmark.
Note that relatively few of the efficiency-optimized networks report results without bells and whistles, and most such results are reported on the development (not test) set of GLUE.
Fortunately, the authors of MobileBERT -- a network which we will find in the next section compares favorably to other efficient networks with bells and whistles enabled -- do provide development-set results without distillation on 4 of the GLUE tasks.\footnote{Note that some papers report results on only the development set or the test set, and some papers only report results on a subset of GLUE tasks. Our aim with this evaluation is to be as inclusive as possible, so we include papers with incomplete GLUE results in our results tables.}
We observe in the upper portion of Table~\ref{T_dev_set} that, when both networks are trained without distillation, SqueezeBERT achieves higher accuracy than MobileBERT on all of these tasks.
This provides initial evidence that the techniques from computer vision that we have adopted can be applied to NLP, and reasonable accuracy can be obtained.
Further, we observe that SqueezeBERT is 4.3x faster than BERT-base, while MobileBERT is 3.0x faster than BERT-base.\footnote{In our measurements, we find MobileBERT takes 572ms to classify one length-128 sequence on a Pixel 3 phone. This is slightly faster than the 620ms reported by the MobileBERT authors in the same setting~\cite{2019_MobileBERT_OpenReview}. We use the faster number in our comparisons. Further, all latencies in our results tables were benchmarked by us.}

Due to the dearth of efficient neural network results on GLUE without bells and whistles, we also provide a comparison in Table~\ref{T_dev_set} with the ALBERT-base network.
ALBERT-base is a version of BERT-base that uses the same weights across multiple attention layers, and it has a smaller encoder than BERT.
Due to these design choices, ALBERT-base has 9x fewer parameters than BERT-base.
However, ALBERT-base and BERT-base have the same number of FLOPs, and we observe in our measurements in Table~\ref{T_dev_set} that ALBERT-base does not offer a speedup over BERT-base on a smartphone.\footnote{However, reducing the number of parameters while retaining a high number of FLOPs can present other advantages, such as faster distributed training~\cite{2019_albert, 2016_firecaffe} and superior energy-efficiency~\cite{2017_small_nn}.}
Further, on the two GLUE tasks where the ALBERT authors reported the accuracy of ALBERT-base, MobileBERT and SqueezeBERT both outperform the accuracy of ALBERT-base.

\vsubsection{Results {\em with} bells and whistles}
\label{sec_results_with_bells}
Now, we turn our attention to comparing SqueezeBERT to other models, all trained with bells-and-whistles.
In the lower portion of Table~\ref{T_test_set}, we first observe that when trained with bells-and-whistles MobileBERT matches or outperforms the accuracy of the other efficient models (except SqueezeBERT) on 8 of the 9 GLUE tasks.
Further, on 4 of the 9 tasks SqueezeBERT outperforms the accuracy of MobileBERT; on 4 of 9 tasks MobileBERT outperforms SqueezeBERT; and on 1 task (WNLI) all models predict the most frequently occurring category.\footnote{Note that data augmentation approaches have been proposed to improve accuracy on WNLI; see \cite{2019_wnli_trick}. For fairness in comparing against our baselines, we choose not to use data augmentation to improve WNLI results.}
Also, SqueezeBERT achieves an average score across all GLUE tasks that is within 0.4 percentage-points of MobileBERT.
Given the speedup of SqueezeBERT over MobileBERT, we think it is reasonable to say that SqueezeBERT and MobileBERT each offer a compelling speed-accuracy tradeoff for NLP inference on mobile devices.

\vsection{Related Work}

{\bf Quantization and Pruning.}
Quantization is a family of techniques which aims to reduce the number of bits required to store each parameter and/or activation in a neural network, while at the same time maintaining the accuracy of that network. This has been successfully applied to NLP in such works as \cite{2019_qbert, 2019_q8bert}. Pruning aims to directly eliminate certain parameters from the network while also maintaining accuracy, thereby reducing the storage and potentially computational cost of that network; for an application of this to NLP, please see~\cite{2020_movement_pruning}. These methods could be applied to SqueezeBERT to yield further efficiency improvements, but quantization and pruning are not a focus of this paper.

{\bf Convolutions in self-attention networks for language-generation tasks.}
In this paper, our experiments focus on natural language understanding (NLU) tasks such as sentence classification.
However, another widely-studied area is natural language generation (NLG), which includes the tasks of machine-translation (e.g., English-to-German) and language modeling (e.g., automated sentence-completion). %
While we are not aware of work that adopts convolutions in self-attention networks for NLU, we {\em are} aware of such work in NLG.
For instance, the Evolved Transformer and Lite Transformer architectures contain self-attention modules and convolutions in separate portions of the network~\cite{2019_evolved_transformer, 2020_lite_transformer}.
Additionally, LightConv shows that well-designed convolutional networks without self-attention produce comparable results to self-attention networks on certain NLG tasks~\cite{2019_pay_less_attention}.
Also, Wang~\etal sparsify the self-attention matrix multiplication using a pattern of nonzeros that is inspired by dilated convolutions~\cite{2020_transformer_on_a_diet}. 
Finally, while not an attention network, Kim applied convolutional networks to NLU several years before the development of multi-head self-attention~\cite{2014_kim_cnn}.

\vsection{Conclusions \& Future Work}

In this paper, we have studied how grouped convolutions, a popular technique in the design of efficient CV neural networks, can be applied to NLP.
First, we showed that the position-wise fully-connected layers of self-attention networks can be implemented with mathematically-equivalent 1D convolutions.
Further, we proposed SqueezeBERT, an efficient NLP model which implements most of the layers of its self-attention encoder with 1D grouped convolutions.
This model yields an appreciable >4x latency decrease over BERT-base when benchmarked on a Pixel 3 phone.
We also successfully applied distillation to improve our approach's accuracy to a level that is competitive with a distillation-trained MobileBERT and with the original version of BERT-base.

We now discuss some possibilities for future work in the directions outlined in this paper. There are several techniques in use in CV that could be applied to NLP which we have not covered in this paper. One very promising direction is downsampling strategies which decrease the sequence length of the activations in the self-attention network as the layers progress. Extensions of this idea, such as U-Nets~\cite{ronneberger2015u}, as well as modifying channel sizes (hidden size) instead of and in addition to sequence length would also be promising directions. On this path of techniques, applying ideas such as BiFPNs~\cite{2020_efficientdet}, striding, and dilation~\cite{2016_yu_dilation} may also yield interesting results.

The addition of all of these potential techniques opens up a significantly broader search-space of neural architecture designs for NLP.
This motivates the application of automated neural architecture search (NAS) approaches such as those described in \cite{2019_squeezenas, 2019_fbnet}.

\clearpage

\section*{Broader Impact}
\label{sec_broader_impact}
\subsection*{Potential benefits of this work}
We hope the techniques used in this paper will allow more efficient and practical deployment of self-attention based networks, particularly allowing more widespread use of these networks on mobile devices. Possible use cases include email sorting, chat analysis, spam detection, or hate-speech filtering.
Facebook has begun using a self-attention based network to automatically detect and remove hate speech on their platform~\cite{dansby_ma_ozertem_stoyanov_yang_fang_moghbel_peng_wang_zhang_et_al, 2020_facebook_hate_speech}. Currently, companies usually run these models on the server-side, but some may be reticent to deploy them on a mass scale due to computation costs. Mobile inference wouldn't require expensive server infrastructure and allows use in cases where privacy and security may be a concern. Running on the edge devices may allow more real-time or offline use cases such as grammar checking, or sentence completion. 

\subsection*{Potential malicious uses of this work}
Given that we are releasing our model and training and inference code as free software, anyone can train our model on any dataset that they like.
While we hope most practitioners will apply our work for altruistic or at least well-intentioned purposes, some may apply our work for ethically questionable or purely self-serving applications.
For instance, low-cost mobile inference could allow "smart" key-loggers and eavesdropping viruses to be deployed, and for these devices to better avoid detection by computing locally and only uploading important information.
And, while we hope that social networks and other text-centric portals will draw on our work to embed fair and just models into their mobile applications for content moderation, these companies could just as easily draw on our work to train and deploy models that censor content of political enemies or amplify only certain types of messages and voices.

\subsection*{Potential effects of unintended bias in the neural network and its training data}
Models for NLP commonly suffer from biases regarding race and gender~\cite{tan2019assessing}.
In our work, we have used standard datasets, and we are not aware of any experimental factors that would increase or decrease SqueezeBERT's propensity for bias, as compared to the approaches used in similar self-attention research such as BERT and ELECTRA.
The gender-related biases of our pretraining corpora (Wikipedia and BooksCorpus) are investigated by Tan~\etal~\cite{tan2019assessing}, and the gender biases of BERT and GPT models finetuned on the GLUE tasks are investigated by Babaeianjelodar~\etal~\cite{2020_quantifying}.

Aside from gender bias, we are not aware of studies that investigate other patterns of bias in the datasets that we used in our work.
However, Sheng~\etal investigate biases regarding race, gender, and sexual orientation in a GPT-2 self-attention model trained on a language-modeling (text-generation) task~\cite{2019_gpt2_bias}.
Further, in the paper introducing the GPT-3 model, a study similar to that of Sheng~\etal is performed~\cite{2020_gpt3}.

According to the taxonomy described in~\cite{2020_language_technology_is_power}, the {\em harms} caused by biased NLP technology include:
\begin{itemize}
   \item Allocational harms: These arise when an "automated system allocates resources (e.g., credit) or opportunities (e.g., jobs) unfairly to different social group."
    \item Representational harms: These arise when "a system (e.g., a search engine) represents some social groups in a less favorable light than others, demeans them, or fails to recognize their existence altogether."
\end{itemize}
We direct the interested reader to work by Blodgett~\etal~\cite{2020_language_technology_is_power} and Sun~\etal~\cite{sun-etal-2019-mitigating} for more details and suggestions on how to minimize the bias in datasets and the neural networks that learn from them.

\begin{ack} %
K. Keutzer's research is supported by Alibaba, Amazon, Google, Facebook, Intel, and Samsung.

\end{ack}

\bibliographystyle{IEEEtran}
\bibliography{bibliography}

\clearpage

{\LARGE \bf
	\begin{center}
		Supplementary Material for SqueezeBERT \\
	\end{center}
}

\appendix %

\setlength{\tabcolsep}{6pt} %

\vsection{Overview of GLUE tasks}
\label{appendix_glue_tasks}
\vspace{-0.2cm}

In this section we provide an overview of each of the tasks within the GLUE benchmark, their evaluation metrics, and their potential applications outside of the benchmark.
For further information on the benchmark, please see~\cite{2018_glue}, where it was originally proposed.

Note that some tasks have one evaluation metric, and some have two.
The possible evaluation metrics are Accuracy (abbreviated below as {\tt acc}), F1~\cite{1979_f_score}, Matthews Correlation Coefficient ({\tt MCC})~\cite{1975_matthews_corr}, Pearson Correlation ({\tt pearson}), and Spearman Correlation ({\tt spearman}).

\textbf{MNLI:} Multi-genre Natural Language Inference~\cite{2018_mnli}. Given a pair of sentences (sentence 1 and sentence 2), the task is to predict whether sentence 2 entails sentence 1. Note that there are 2 test sets for this task: MNLI-matched (MNLI-m), and MNLI-mismatched (MNLI-mm).
When computing an average score overall tasks, we follow the convention employed by the GLUE leaderboard, which is {\tt (acc(MNLI-m) + acc(MNLI-mm))/2}.
Therefore, while MNLI-m and MNLI-mm are two columns in our results table, they only count as one task in the average GLUE score. Potential applications include helping users improve their writing by checking for repetitive sections based on whether certain sentences are entailed by other previous ones. This task contains 392,703 training examples.

\textbf{QQP}: Quora Question Pairs~\cite{2017_qqp,2018_qqp}. The goal of this task is to determine whether a pair of questions have the same meaning. In practical applications, such as organizing emails or organizing helpdesk tickets, this approach can be used to group similar questions together and only answer them once.
There are 2 metrics for this task: {\tt acc} and {\tt F1}.
In our results tables, we follow the approach of the GLUE leaderboard and compute QQP results as the average of these two metrics.
We observe that, for most models, on the test set the F1 score is significantly lower than the accuracy score.
For instance, previously reported BERT-base results are F1=71.2 and accuracy=89.2.
The TinyBERT paper~\cite{2019_tinybert} reports a TinyBERT QQP score of 71.3, relative to a BERT-base baseline of 71.2, so it appears that the TinyBERT paper only reports the QQP F1 (but not accuracy) score.
Conversely, the ELECTRA~\cite{2020_electra} paper reports an ELECTRA-small QQP score of 88.0, relative to a BERT-base QQP score of 89.2, so it appears that the ELECTRA paper only reports the QQP accuracy (but not F1) score.
Out of fairness, we have omitted the TinyBERT and ELECTRA-small QQP test-set results from our results table, as it appears that they are each using a metric different from our metric of ({\tt (acc(QQP) + F1(QQP))/2}). This task contains 363,871 training examples.

\textbf{QNLI}: Question-answering Natural Language Inference. This task is derived from the dataset published in \cite{2016_squad}. Specifically, it is reformulated as a two-class classification task, wherein the goal is to decide whether or not the given sentence does, or does not, contain the answer to a given question. In this way, QNLI bears some similarity to other entailment classification tasks such as MNLI. The evaluation metric used for QNLI is just {\tt acc(QNLI)}. In terms of applications, QNLI is useful for many of the same situations in which a standard question-answering dataset would be, such as in designing intelligent voice assistants which can respond to user questions. For instance, if an extractive or generative model is used to generate an answer to a question, a QNLI-trained model could be used as a sanity-check on whether the machine-generated response actually answers the question. Also, a QNLI-trained model could be used to check if an existing answer already answers a new question on a form or in a helpdesk database, thereby not requiring a new answer to be generated by a human. This task contains 104,744 training examples.

\textbf{SST-2}: Stanford Sentiment Treebank \cite{2013_sst2}. This is a sentiment classification dataset, with the goal being to predict whether a sentence has a positive or negative sentiment. The source data for SST-2 is hand-annotated movie review data. The metric used for this task is {\tt acc(SST-2)}. SST-2 applications include improving chatbot based systems which require responses that properly match the conversation context to be generated, and potentially triaging emails at helpdesks so that, for example, the ones with strong negative sentiments are prioritized. This task contains 67,350 training examples.

\textbf{CoLA}: Corpus of Linguistic Acceptability \cite{2019_cola}. This is a binary classification task, where the objective is to determine whether a given sentence meets the criteria for being a properly written English sentence. It is worth noting that CoLA is a binary classification task, and 69\% of the data samples in the validation set are positive examples. The metric used for this task is different from the metrics used in other GLUE tasks: {\tt MCC(CoLA)}. 
A distinguishing feature of Matthews Correlation Coefficient ({\tt MCC}) is that it effectively adjusts for class imbalance in the dataset.
On the CoLA validation set, where 69\% of the samples are positive, predicting the majority class would produce an accuracy of 69\% but an {\tt MCC} of 0.0.
Also note that {\tt MCC} can be anywhere between and including -1.0 and 1.0, as opposed to 0.0 (0\%) and 1.0 (100\%) for {\tt acc}. Models trained on CoLA could be used to provide feedback on the grammatical correctness of user-generated text, analogous to a spell-checker, but potentially with more nuanced understanding of grammar than a rule-based grammar checker. CoLA-trained models could also be used to verify the grammatical correctness of text generated by other neural networks. This task contains 8,551 training examples.

\textbf{STS-B}: Semantic Textual Similarity Benchmark \cite{2017_stsb}. This dataset is built from several sources including news headlines data as well as captions data. The goal of this task is to predict the level of semantic similarity between two input sentences, on a scale of 1 (minimum similarity) to 5 (maximum similarity). Notably, this is a regression task. The overall evaluation metric used on the GLUE leaderboard, which we report in our results tables, is {\tt (spearman(STS-B) + pearson(STS-B))/2}. STS-B could have applications in several areas, such as plagiarism detection, as well as potentially other applications which are mentioned above in the description of QQP. This task contains 5,750 training examples.

\textbf{MRPC}: Microsoft Research Paraphrase Corpus \cite{2005_mrpc}. This is classification task where the goal is to predict whether or not the two input sentences have the same semantic meaning. The data itself comes from online news articles and is hand-annotated. The metric used for this task is ({\tt (acc(QQP) + F1(QQP))/2}), consistent with the GLUE test set leaderboard. MRPC would have applications similar to STS-B, such as plagiarism detection and others. This task contains 3,669 training examples.

\textbf{RTE}: Recognizing Textual Entailment is a combination of several datasets from yearly challenges on entailment \cite{2005_rte1, 2006_rte2, 2007_rte3, 2009_rte5}. RTE is a two-way classification task where the labels are \textit{entailment} and \textit{not\_entailment}. The metric used for this task is {\tt acc(RTE)}. Potential applications of this dataset are similar to those of MNLI. This task contains 2,491 training examples.

\textbf{WNLI}: Winograd Schema Challenge, reformulated as Winograd Natural Language Inference \cite{2012_wnli}. This goal of this task is for the model to predict the antecedent of a pronoun (out of a set of choices) given the input sentence containing that pronoun. The metric used for this task is {\tt acc(WNLI)}. As with other work in this area \cite{2019_BERT}, we simply predict the majority class on this dataset, resulting in a test set accuracy of 65.1\%. This task contains 636 training examples.

In Listings~\ref{lst_dev_score} and~\ref{lst_test_score}, we show how we calculate overall scores for the GLUE development set and test set.
Other than omitting WNLI from the development-set score (following~\cite{2019_BERT}), these calculations adhere to the methodology used by the GLUE leaderboard.

\begin{lstlisting}[language={Python},
                   caption={Calculating average score on the GLUE {\bf development} set.},
                   captionpos=t,
                   label={lst_dev_score}]
dev_score = MEAN(
                 (acc(MNLI-m) + acc(MNLI-mm))/2,
                 (F1(QQP) + acc(QQP))/2,
                 acc(QNLI),
                 acc(SST-2),
                 MCC(CoLA),
                 (pearson(STS-B) + spearman(STS-B))/2,
                 (F1(MRPC) + acc(MRPC))/2,
                 acc(RTE)
                )
\end{lstlisting}

\begin{lstlisting}[language={Python},
                   caption={Calculating average score on the GLUE {\bf test} set.},
                   captionpos=t,
                   label={lst_test_score}]
test_score = MEAN(
                  (acc(MNLI-m) + acc(MNLI-mm))/2,
                  (F1(QQP) + acc(QQP))/2,
                  acc(QNLI),
                  acc(SST-2),
                  MCC(CoLA),
                  (pearson(STS-B) + spearman(STS-B))/2,
                  (F1(MRPC) + acc(MRPC))/2,
                  acc(RTE),
                  acc(WNLI)
                 )
\end{lstlisting}

\vsection{More training details and results}

\vsubsection{Training Hardware}
We do all pretraining and finetuning on an 8-GPU server without multi-server distributed training.
Our server has 8 NVIDIA Titan RTX GPUs.
The server also has an Intel Xeon Gold 6130 64-core CPU, and it has 256GB of RAM.
Further, the data loading requirements for our NLP application are so small that we were able to use low-cost, low-bandwidth spinning-media (not flash) disks for storing the training data.
We employed two of these servers for approximately 6 months to do various experiments, beginning with reproducing BERT-base, and then experimenting with various model architectures.
No cloud computing resources or corporate computing resources were used for this research.

While we used our own computing resources for this work, we now consider what it {\em would have cost} to rent servers similar to ours from a cloud provider for the duration of our project.
We can break this down into two types of costs, {\em storage} and {\em computation}:
\begin{itemize}
    \item {\bf Storage.} The datasets are small (under 100GB total), and we typically retain a few hundred gigabytes of model parameters from various training checkpoints, so we imagine the cloud storage costs would be negligible: As of this writing, Amazon Web Services charges \$0.023/GB for general-purpose storage~\cite{2020_aws_s3}, so storing a terabyte of data and model checkpoints would cost just \$23 per month, or \$138 for 6 months. There may be additional fees for data transfers within the AWS ecosystem, but we have studied these costs in detail.
    
    \item {\bf Computation.} As we mentioned earlier, to do the work in this project, including the initial work to reproduce baselines and the work to experiment with several potential neural network designs, we used 2 8-GPU servers for about 6 months. We estimate that renting this amount of computation from a cloud provider would cost around \$100,000, calculated as \\ {\footnotesize (6 months) * (50\% utilization) * (2 servers) * (30 days per month) * (24 hrs/day) * (\$24/hr for an 8-GPU V100 machine from Amazon Web Services~\cite{2020_aws_p3})}.
\end{itemize}
So, while we used our computing resources for this work, doing this entire research project from beginning to end in the cloud would have cost around \$100,000.

Now, we consider what it would cost to reproduce the SqueezeBERT model from scratch using cloud hardware.
The SqueezeBERT model can be reproduced in approximately 5 days: 4 days for pretraining, and then under one day for finetuning on all GLUE tasks with the optimal hyperparameters discovered by our hyperparameter search (see Section~\ref{sec_hparam_search}).
In total, this would cost approximately \$2880, calculated as (5 days) * (24 hrs/day) * (\$24/hr for one 8-V100 AWS machine).

\vsubsection{Training Software}
Our PyTorch-based training and inference code draws heavily on the HuggingFace Transformers~\cite{2019_huggingface_transformers} and NVIDIA Deep Learning Examples~\cite{2020_deeplearningexamples} repositories.
We perform pretraining using 8 GPUs with 16-bit floating-point math, and we use the O2 optimization level in the NVIDIA Apex mixed-precision training primitives~\cite{2020_apex}.
We perform finetuning on a single GPU with 32-bit floating-point math, and we concurrently run multiple finetuning tasks across the 8-GPU machine.

\vsubsection{Further details: distillation}
In the distillation approach that we described in Section~\ref{sec_training_with_bells}, we mentioned that we use teacher logits and one-hot ground-truth as the target output for our soft cross entropy loss.
The weighting between the teacher logits and the ground-truth is controlled by a hyperparameter $\alpha$.
Let $\Psi_t$ represent the teacher logits and let $\Psi_g$ represent the one-hot encoding of the ground-truth.
Formally, we write this weighted sum as:
\begin{equation}
    \label{eq_weighted_sum}
    \Psi = (1-\alpha) \Psi_t + \alpha \Psi_g
\end{equation}

In the next section, we will tune $\alpha$ as part of our hyperparameter tuning scheme.

\vsubsection{Details of hyperparameter search during finetuning on GLUE tasks}
\label{sec_hparam_search}
We now present more details on the hyperparameter search approach that we used for training SqueezeBERT with bells and whistles.
In Table~\ref{T_hparam_space}, we present the space of possible hyperparameters over which we performed a grid-search.
Note that the time to finetune the model using one set of hyperparameters varies significantly depending on the GLUE task, from 15 minutes for small datasets like RTE, to 14 hours for MNLI.
For smaller datasets that require less training (e.g. RTE and MRPC), we use a broader search space and more epochs.
And, for larger datasets (e.g. MNLI and QQP), we use a more narrow search space with fewer epochs.

Now, in Table~\ref{T_hparam_results}, we present the best hyperparameters found in our search.
We observe two interesting phenomena on this table.
The first is regarding the use of distillation.
Recall from Equation~\ref{eq_weighted_sum} that $\alpha$ is the hyperparameter that sets the weighting between the teacher logits and the ground-truth for distillation.
When $\alpha=1.0$, the teacher logits are ignored, and thus distillation is disabled.
So, it is interesting to note that on three of the eight GLUE tasks in Table~\ref{T_hparam_results} (MNLI, QNLI, and CoLA), distillation did not produce superior results over non-distillation finetuning.
The second interesting phenomenon in this table is that maximum accuracy was not necessarily achieved on final epoch.
For example, on QNLI, SqueezeBERT converged to its maximum development-set accuracy after just two epochs.

\vsection{Further details: Inference on a smartphone}
To evaluate inference speed, we run the neural networks on a Google Pixel 3 smartphone.
This phone contains the popular Qualcomm Snapdragon 845 processing chip, which is also used in popular smartphones from Samsung, Xiaomi, and Sony~\cite{2019_snapdragon_845_list}.
The phone also has 4GB of LPDDR4x memory.
To run a neural network on the phone, we do the following.
First, we export the network to TorchScript using the {\tt torch.jit.trace()} command, which yields a standalone neural net that does not require Python to run.
Then, we copy the network to the phone, and we run it using a modified version of the {\tt speed\_benchmark\_torch.cc} file that is built into PyTorch.
We report the average latency of 40 runs, using the CPU cores of the phone.
Following the protocol of MobileBERT, each run operates on a length-128 input sequence and a batch size of 1.

\begin{table*}[t!]
    \caption{Our hyperparameter search space.}
    \centering
    \footnotesize
    \begin{tabular}{l|lll}
    \toprule
      Hyperparameter & MNLI, QQP, QNLI, STS-2 & STS-B, MRPC, RTE & CoLA \\
      \midrule
      $\alpha$ & [0.8, 0.9, 1.0]  & [0.8, 0.9, 1.0] & [0.8, 0.9, 1.0] \\
      Learning Rate & {\tiny [1e-05, 2e-05, 3e-05, 4e-05]} & {\tiny [1e-05, 2e-05, 3e-05, 4e-05, 5e-05]} & {\tiny [1e-05, 2e-05, 3e-05, 4e-05, 5e-05]} \\
      Encoder Dropout & [0.0, 0.1] & [0.0, 0.1] & [0.0, 0.1]  \\
      Final Dropout & [0.1, 0.2] & [0.1, 0.2] & [0.0, 0.1]\\
      Epochs & 5 & 10 & 20 \\
      Batch Size & 16 & 16 & [16, 32, 48] \\
      \bottomrule
    \end{tabular}
    \label{T_hparam_space}
\end{table*}

\setlength{\tabcolsep}{5pt} %
\begin{table*}[t!]
    \caption{Hyperparameters selected by our hyperparameter search when training SqueezeBERT with bells-and-whistles.}
    \centering
    \footnotesize
    \begin{tabular}{l|lllllllll}
    \toprule
      Hyperparameter & MNLI-m & MNLI-mm & QQP & QNLI & STS-2 & CoLA & STS-B & MRPC & RTE \\
      \midrule
      $\alpha$ & 1.0 & 1.0 & 0.8 & 1.0 & 0.8 & 1.0 & 0.8 & 0.9 & 0.8 \\
      Learning Rate & {\tiny 3e-05} & {\tiny 3e-05} & {\tiny 4e-05} & {\tiny 3e-05} & {\tiny 3e-05} & {\tiny 2e-05} & {\tiny 4e-05} & {\tiny 3e-05} & {\tiny 3e-05}\\
      Encoder Dropout & 0.0 & 0.0 & 0.1 & 0.0 & 0.1 & 0.0 & 0.1 & 0.1 & 0.1 \\
      Final Dropout   & 0.1 & 0.1 & 0.1 & 0.1 & 0.1 & 0.1 & 0.1 & 0.1 & 0.1 \\
      Epochs & 4 & 4 & 5 & 2 & 5 & 5 & 10 & 9 & 3 \\
      Batch Size & 16 & 16 & 16 & 16 & 16 & 16 & 16 & 16 & 16 \\
      \bottomrule
    \end{tabular}
    \label{T_hparam_results}
\end{table*}

\vsection{More related work}
{\bf Efficient convolutional networks for CV.}
Convolutional neural networks lead the state-of-the-art on computer vision (CV) tasks such as image classification, object detection, and semantic segmentation.
In the last five years, the community has developed efficient convolutional neural networks for CV that run efficiently and in real-time on mobile devices.
On the ImageNet~\cite{2015_imagenet} image classification task, from the year 2016 (ResNet-101~\cite{2015_resnet}) to 2020 (FixEfficientNet-D0~\cite{2020_fixefficientnet}), there is a 20x reduction in the number of floating-point operations (FLOPs) required, while accuracy has actually improved.\footnote{In our terminology throughout the paper, a multiply-add operation is two FLOPs.}\footnote{These results hold without expanding the training data beyond the ImageNet training set.}
Further, on the COCO~\cite{2014_coco} object detection task, from 2016 (FPN~\cite{2017_fpn}) to 2020 (EfficientDet-D1~\cite{2020_efficientdet}), there is a 38x reduction in the number of FLOPs and an improvement in accuracy.\footnote{FPN and EfficientDet-D1 were both pretrained on COCO and finetuned on ImageNet.}
Finally, on the Cityscapes~\cite{2016_cityscapes} semantic segmentation task, from 2016 (FCN-8s~\cite{2015_fcn}) to 2020 (SqueezeNAS-MAC-Small~\cite{2019_squeezenas}), there is a 334x reduction in the number of FLOPs, and an improvement in accuracy.\footnote{FCN-8s was pretrained on ImageNet and finetuned on Cityscapes. SqueezeNAS-MAC-Small was pretrained on ImageNet and COCO and finetuned on Cityscapes. Chen~\etal found that pretraining on COCO improved Cityscapes accuracy by approximately 2 percentage-points~\cite{2018_deeplabv3_plus}.}
Given that the datasets were, for the most part, held constant, to what do we owe these improvements?
One factor that has helped is the design of new neural network architectures with superior FLOP-accuracy tradeoffs.
There have been several innovations to neural architecture for CV, including dilated convolutions~\cite{2016_yu_dilation}, creative approaches for multiscale recognition~\cite{2017_fpn, 2020_efficientdet}, and long-range aggregation of skip-connections~\cite{2017_densenet}.
However, above all else, there is one influential design element that has been adopted by nearly all efficient convolutional neural network designs developed in the last three years: grouped convolutions.
Grouped convolutions, which are form of structured sparsity in the channel dimension, have a hyperparameter $G$, which reduces the number of parameters and the number of computations by $\frac{1}{G}$
Grouped convolutions were proposed by Krizhevsky \etal in 2012~\cite{2011_cuda_convnet, 2012_alexnet}, but they were largely forgotten, and they were not used by subsequent state-of-the-art networks such as VGG~\cite{2014_vgg}, Inception~\cite{2014_inception}, and ResNet~\cite{2015_resnet}.
Grouped convolutions began to reemerge in the literature in 2016 with Xception~\cite{2016_xception} and ResNext~\cite{2016_resnext}.
For ImageNet image classification, MobileNet showed in 2017 that in certain cases grouped convolutions can preserve most of the accuracy (compared to a non-grouped baseline) while producing extreme reductions in FLOPs, particularly when setting $G=C$ (where $C$ is the number of channels) in certain layers~\cite{2017_mobilenets}.\footnote{The special case of grouped convolution with $G=C$ is known as a depthwise convolution.}
The efficient FixEfficientNet, EfficientDet, and SqueezeNAS networks that we covered above all make extensive use of grouped convolutions.

{\bf Improved training regimes.}
A number of techniques have been developed to train a given self-attention neural network to higher accuracy on sentence classification tasks.
GPT and BERT proposed methods for self-supervised pretraining of attention networks on large corpora such as Wikipedia to improve sentence classification accuracy~\cite{2018_gpt,2019_BERT}.
STILTS showed that, after performing the BERT pretraining scheme, accuracy can be further improved by applying transfer learning across multiple sentence classification datasets~\cite{2018_stilts}.
Further, MT-DNN showed that training BERT to simultaneously perform multiple sentence classification tasks can yield higher accuracy on some tasks~\cite{2019_mtdnn}.\footnote{When running sentence classification tasks (e.g sentiment analysis {\em and} linguistic correctness checking), the MT-DNN~\cite{2019_mtdnn} approach amoritizes much of the neural network computation across multiple tasks. We are interested to explore this angle for potential further speedups in future work.}
In addition, RoBERTa showed that pretraining BERT for more iterations on a larger dataset yields higher accuracy on NLU tasks~\cite{2019_roberta}.
Further, ELECTRA trained BERT using an adversarial generator-discriminator method, and it achieved superior accuracy to RoBERTa without changing the design of the BERT encoder~\cite{2020_electra}.
The RoBERTa and ELECTRA training regimes yield significant accuracy improvements when applied to both BERT-base and BERT-large, which suggests that these regimes improve the accuracy of larger (higher latency) as well as smaller (lower latency) network architectures.
Finally, we believe it may be possible to further improve the accuracy of SqueezeBERT by pretraining on more data and for more iterations (similar to RoBERTa) and by pretraining with an adversarial method such as ELECTRA.

\end{document}